\documentclass[a4paper,twoside]{article}

\usepackage{epsfig}
\usepackage{subcaption}
\usepackage{calc}
\usepackage{amssymb}
\usepackage{amstext}
\usepackage{amsmath}
\usepackage{amsthm}
\usepackage{multicol}
\usepackage{pslatex}
\usepackage{apalike}
\usepackage{subcaption}
\usepackage{url}
\usepackage{pifont}
\usepackage{adjustbox}
\usepackage{SCITEPRESS}     

\usepackage{multirow} 

\begin{document}

\title{Towards Deep Learning-based 6D Bin Pose Estimation in 3D Scans}

\author{\authorname{paper 190}}

\author{\authorname{Lukáš Gajdošech\sup{1,2,}\orcidAuthor{0000-0002-8646-2147}, Viktor Kocur\sup{2,4,}\orcidAuthor{0000-0001-8752-2685
}, Martin Stuchlík\sup{1,}\orcidAuthor{0000-0001-8556-8364}, Lukáš Hudec\sup{3,}\orcidAuthor{0000-0002-1659-0362}, Martin Madaras\sup{1,2,}\orcidAuthor{0000-0003-3917-4510}}
\affiliation{\sup{1}Skeletex Research, Slovakia}
\affiliation{\sup{2}Faculty of Mathematics, Physics and Informatics, Comenius University Bratislava, Slovakia}
\affiliation{\sup{3}Faculty of Informatics and Information Technologies, Slovak Technical University Bratislava, Slovakia}
\affiliation{\sup{4}Faculty of Information Technology, Brno University of Technology, Czech Republic}
\email{\{gajdosech, kocur\}@fmph.uniba.sk, \{madaras, stuchlik\}@skeletex.xyz}
}



\abstract{An automated robotic system needs to be as robust as possible and fail-safe in general while having relatively high precision and repeatability. Although deep learning-based methods are becoming research standard on how to approach 3D scan and image processing tasks, the industry standard for processing this data is still analytically-based. Our paper claims that analytical methods are less robust and harder for testing, updating, and maintaining. This paper focuses on a specific task of 6D pose estimation of a bin in 3D scans. Therefore, we present a high-quality dataset composed of synthetic data and real scans captured by a structured-light scanner with precise annotations. 
Additionally, we propose two different methods for 6D bin pose estimation, an analytical method as the industrial standard and a baseline data-driven method. Both approaches are cross-evaluated, and our experiments show that augmenting the training on real scans with synthetic data improves our proposed data-driven neural model. This position paper is preliminary, as proposed methods are trained and evaluated on a relatively small initial dataset which we plan to extend in the future.
}

\onecolumn \maketitle \normalsize \setcounter{footnote}{0} \vfill

\section{\uppercase{Introduction}}
\label{sec:introduction}

Capturing a scene with 3D scanners is a standard for automatized systems analyzing a scene. To pick mechanical parts from a bin by a robotic arm equipped with a gripper, the parts need to be localized. 
First, the localization of bin is essential to restrain the robot from collisions. Then, the kinematics of the robot is optimized for path planning.
The problem of bin localization can be defined as a 6 DoF pose estimation of a template 3D model of the bin in the 3D scan. 

Nowadays, analytical methods are still the industrial standard for the processing of 3D scans. On the contrary, the academic and research standards have evolved to data-driven or hybrid approaches. Analytical computation of bin transformation in captured point clouds might be vulnerable to missing critical information in the captured scans, like corners and edges, yielding lower robustness than expected. 
The computation precision of a hard-defined analytical algorithm might be higher but at the cost of lower robustness if a key content is missing. In applications of automated intelligent systems, it may be interesting to lower its precision to increase the robustness in some scenarios. The other possible approach is to split the pipeline into two steps - the first part of the pipeline orients on the robustness and raw data-driven localization. The second part focuses on the precision-based analytical solution starting from the predicted pose estimations, thus having the robustness properties inherited from the data-driven approach.

In this paper, we present a novel dataset containing high-quality real and synthetic 3D scans of different bins in various poses containing a variety of items captured by structured light scanners. We publish the dataset\footnote{\url{http://skeletex.xyz/portfolio/datasets}} for further research. We propose an analytical method and a conceptually simple deep convolutional neural network for 6D bin pose estimation. We experimentally evaluate it and show that our network is more robust than the analytical method. Our method achieves better accuracy than existing 6D pose estimation methods. We also show the inclusion of synthetic data into the training process is beneficial. We experimentally verify that cases of successful pose approximations done by our network can be further refined in post-processing with iterative closest point (ICP), substantially increasing the portion of data with close-to-zero final error. 
We present this work as a position paper.
We nevertheless feel that the preliminary results presented in this paper show promise, and we intend to continue this research by collecting a larger dataset and performing a more thorough evaluation.


\section{\uppercase{Related work}}
Finding the 6D pose of an object is one of the classical computer vision problems tackled using various methods over the years. Existing algorithms for images and point clouds categorize into two main groups, analytical algorithms \cite{stein:1992:structural,katsoulas:2003:hough1}, and data-driven algorithms. Data-driven algorithms can be further split into feature-based methods \cite{vidal:2018:sensors,drost:2010:model} and Deep Neural Networks (DNN)-based methods \cite{park:2019:pix2pose,bukschat:2020:efficientpose}. 

On the one hand, the feature-based methods are optimized using only the 3D object model, as they match pairs of points between the model and the captured scene. DNN-based methods, on the other hand, are trained on large sets of actual 3D scenes to generalize the solution. Moreover, a hybrid method can be composed of a sequence of data-driven steps and the final analytical step, with the ICP-like methods being the widely-used analytical post-processing step \cite{besl:1992:icp,xiang:2017:posecnn}. BOP challenge is trying to capture the state-of-the-art in this area, comparing traditional and data-driven methods on benchmark datasets \cite{hodan:2020:bop}.

Even though the problem of finding 3D translation and 3D rotation of rigid objects is very general, it is nevertheless dependant on the input data. Most of the widely adopted datasets consist of RGBD images of textured objects with complex geometries from a single device with known internal camera parameters. 

\subsection{Analytical and Feature-based Methods}
A traditional approach of registering objects has been detecting the local descriptors combined into shape-based primitives and searching for their corresponding pairs on 3D CAD models. The simplest case is Hough transform applied to detect lines \cite{katsoulas:2003:hough1}. The efforts to enhance the algorithm to reduce the number of possible detections resulted in specifying that lines have to be orthogonal to represent the shape borders \cite{katsoulas2003localization}. 

Similar to Hough transform, the RANSAC algorithm extracts the geometric description of the object by fitting the corresponding shape primitives into the 3D data. The non-deterministic algorithm is used in a sequence of standard steps. \cite{GUO2020} enhance the algorithm by using shape primitives to approximate the objects. \cite{VOCK201936} propose to reduce 3D points into point pair features (PPF). However, RANSAC usually ends with many false positives (e.g., floor points); therefore, an ICP is usually required for fine-tuning. PPFs are widely used in literature to estimate object points in point cloud or RGBD data \cite{drost:2010:model,vidal:2018:sensors,guo2021efficient}. 


\subsection{Deep Neural Network-based Methods}

Some methods estimate 6D poses from a single RGB image either directly by modifying an existing 2D object detection framework \cite{bukschat:2020:efficientpose} or by using a neural network to obtain 2D-3D correspondences further used in a PnP solver to obtain the final pose \cite{park:2019:pix2pose,zakharov:2019:dpod}. 
In contrast to RGB data, the scanners utilized in our work output only texture from a grayscale camera - not color, limiting the application of related papers even further.

RGB with depth information is also commonly used as input for deep learning base pose estimation. Several methods \cite{mitash2018,jafari2019} use deep learning models to output hypotheses which are then processed in a hypothesis validation pipeline to obtain the final poses. Other indirect methods use deep learning networks to output keypoints \cite{he:2020:pvn3d} or object fragments \cite{hodan:2020:epos} which are then used in a PnP solver to obtain the final poses.

Other deep learning approaches apply neural networks directly to compute the 6D pose. DenseFusion \cite{wang:2019:densefusion} uses RGB information to obtain segmentation masks of objects. These are used to combine depth and RGB data to generate per-pixel embeddings, which are then used to estimate object poses in a voting scheme. An improved version of the algorithm called MaskedFusion \cite{pereira:2020:maskedfusion} improves accuracy by masking non-relevant data.

These approaches are trained for specific objects and require their 3D models to be available during training. We aim to be able to estimate 6D poses of arbitrary bin-shaped objects. The mentioned methods are thus not easily transferable to our scenario. Moreover, the methods are usually trained for cameras with specific internal parameters, a constraint we aim to avoid in our work.

\subsection{Pose Parameterization}

\label{sec:rel:param}

The pose of a rigid object can be described with a pair of a rotation matrix $R \in SO(3)$ and a translation vector $\vec{t} \in \mathbb{R}^3$. The translation vector can usually be represented directly as an output of a neural network and used in a loss function since the space $\mathbb{R}^3$ has a direct continuous representation. On the other hand, there are no continuous representations of $SO(3)$, making it difficult for neural networks to learn such representations \cite{zhou:2019:rotation}. 


\begin{figure}[t]
  \centering
  \includegraphics[width=1\columnwidth]{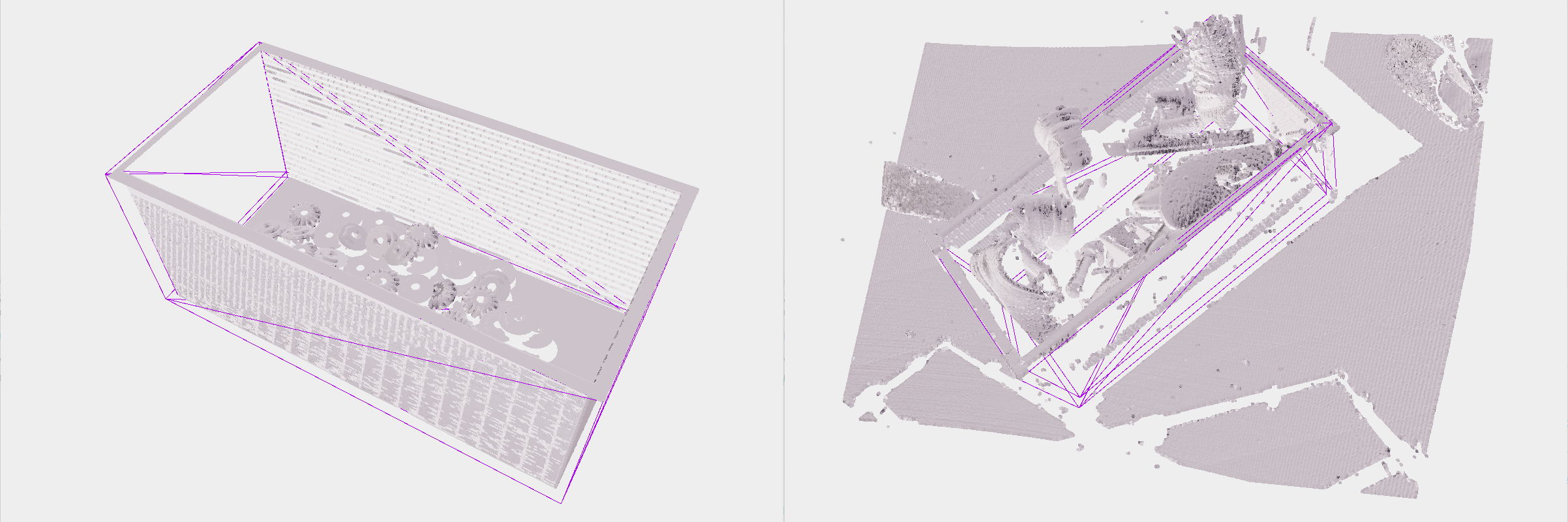}
  \caption{\label{fig:binloca}%
    In this work we present a novel dataset containing 520 real and 370 synthetic 3D scans of bins. (Left) Synthetic sample. (Right) Real scan annotated by hand. The ground truth transformations of bin 3D model into the scanner-space is demonstrated by purple mesh.}
\end{figure}

Rotational matrices only have 3 degrees of freedom while having 9 elements. Constraining the elements directly during the training process is impractical, so an orthogonalization procedure must be utilized \cite{zhou:2019:rotation}. Rotation can also be represented using different equivalent parameterizations such as quaternions \cite{xiang:2017:posecnn,wang:2019:densefusion,pereira:2020:maskedfusion} or axis-angle vectors \cite{bukschat:2020:efficientpose}. 

Symmetric objects pose a specific problem for rotation representation. Depending on the type of symmetry, multiple different rotation parameterizations can be valid for the same pose. This might introduce problems as some loss functions can then have undesirable multiple global minima. Some approaches mitigate these issues for some types of symmetries by using losses based on distances of sampled points on object models \cite{xiang:2017:posecnn,wang:2019:densefusion,pereira:2020:maskedfusion} 
A different approach \cite{pitteri:2019:symmetries} proposes mapping all representations onto a single canonical representation used during training. Some methods avoid these issues altogether by not directly outputting the object pose but calculating it indirectly from keypoints \cite{he:2020:pvn3d} or object fragments \cite{hodan:2020:epos}.

\section{\uppercase{Dataset}}
\label{section:dataset}

We have collected a new dataset consisting of both real captures (scans) from Photoneo PhoXi structured light scanner devices \cite{Photoneo:2019} annotated by hand and synthetic samples produced by our generator. See Figure \ref{fig:binloca} for an example of both real and synthetic 3D scanner captures of scenes composed of mechanical parts in a bin from our dataset.

In comparison with existing datasets, some notable differences include:
\begin{itemize}
    \item most of the captured bins are texture-less, made from uniform, single-colored materials,
    \item all bins are of cuboid shape with different proportions. Compared to objects with complex geometry, bins consist of flat faces with edges, which are not guaranteed to be seen in the capture due to occlusion. Surface models of these bins are not provided, just their approximate bounding boxes,
    \item PhoXi scanner provides high-resolution 3D geometry data, but no RGB data, with a rough and noisy gray-scale intensity image being the closest equivalent,
    \item captures come from different devices with various intrinsic camera parameters. We aim to work directly on 3D point clouds, which contain these parameters implicitly as opposed to RGBD images.
\end{itemize}

The original scans contain various parameters, such as gray-scale intensities and normals. We rely only on 2D single-view maps of 3D coordinates in $2064 \times 1544$ resolution in our proposed approaches. We use 80\% as the training data, and the remaining 20\% (every fifth sample) plus a unique set of independently captured 49 samples (including 10 synthetic samples) as a test set. Due to its currently limited size, we recommend cross-validation instead of an explicit train-validation split. We plan to add more samples into the dataset, as we will further enhance our methods in the future.
 

\section{\uppercase{Bin Pose Estimation in 3D Point Cloud}}

The bin pose estimation is a computation process of estimating a transformation matrix that maps coordinates of a bin-space into a scanner-space. As outlined in the previous sections, the specific task of bin pose estimation differs in many key aspects from the general task of 6D pose estimation. Therefore, we have decided to propose also two methods for this task. The first method is an analytical heuristic we have developed, and the second is a CNN-based pose estimation method. We deliberately designed the methods to be conceptually simple to provide solid baselines without bells and whistles. The following sub-sections describe the proposed methods. Evaluation and comparison of results for a set of experiments are in Section \ref{sec:experiments}. 

\subsection{Analytical Edge-based Fitting}
\label{sec:analytical}

An analytical algorithm for pose estimation is composed of a set of steps performed sequentially in the pipeline. This four-step method assumes that the top edges of the bin are closer to the camera than background objects, and at least a part of every top edge can be seen.

\begin{figure}[ht]
  \centering
  \includegraphics[width=.99\linewidth]{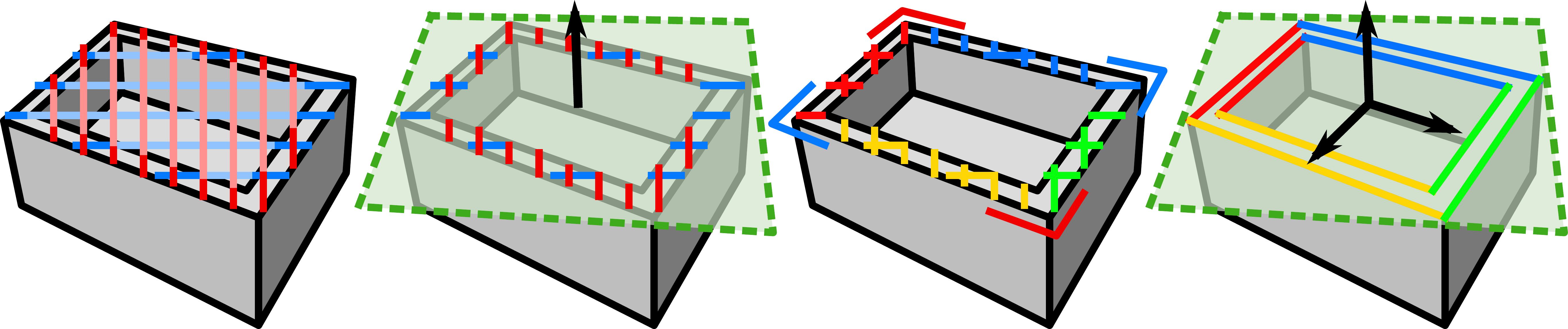}
   \caption{\label{fig:bin_directions}%
            (From left to right) the camera space is row-wise and column-wise segmented into similar depth intervals, from which horizontal and vertical bin-cuts are constructed. A plane is fitted into the bin-cuts, and wall-cuts not corresponding to this plane are discarded as outliers. The remaining wall-cuts are assigned to four bin walls according to corners fitted into horizontal and vertical bin-cuts. Finally, the lines are fitted into categorized wall-cuts, which define the bin basis.}
\end{figure}

Initially, the horizontal and vertical scan-lines are defined in scan-space. Each scan-line is divided into intervals, where scan-line interval going through the whole bin is called bin-cut. Specifically, each bin-cut is composed of two wall-cuts and one interval for the floor (representing the ground of the bin). 

Next, minimum depth values in camera-space in the intervals are detected, and vectors describing edge-to-edge direction are computed. The set of such vectors is computed in both directions, horizontally and vertically (see Figure \ref{fig:bin_directions}, left).

Moreover, a mode vector direction is computed in both horizontal and vertical directions. Those mode directions are used to compute the cross product of these directions to compute the normal defining the top of the bin. At the end of the step, the wall-cuts are filtered according to the calculated plane. 

Consequently, a corner detection is performed on the filtered data. Each corner is detected as a bin-cut endpoint, where the change of direction between neighboring bin-cut endpoints is the highest; such detection is performed in every direction, and all four corners are detected (see Figure \ref{fig:bin_directions}, right). 

Finally, the set of detected corners categorizes wall-cuts into four categories of the bin walls. Lines are fitted into filtered wall-cuts, and the bin-space is defined using the computed plane normal and fitted lines, which can be used for the bin-space definition and calculation of the final bin-space to camera-space transformation.

\subsection{CNN-based Pose Estimation}
\label{sec:neural}

The analytical method may fail when bin edges or corners are occluded or outside of the scanner view. Such instances may frequently occur in industrial applications when human or robotic operators manipulate bins or contain items that cover the bin edges.

To overcome these issues, we propose a data-driven approach using a convolutional neural network. We propose a simple network that can reliably estimate the pose up to a reasonable level of accuracy. This estimate provided by the network is then refined using an ICP algorithm to obtain the final bin pose. 

\subsubsection{Parameterization of the bin pose}

The pose of the bin can be parameterized using a rotation matrix $R \in SO(3)$ and a translation vector $\vec{t} \in \mathbb{R}^3$. 
We represent the translation vector directly. To represent rotation, we opt to use a strategy similar to \cite{zhou:2019:rotation} and represent the rotation by using two vectors from $\mathbb{R}^3$ which can be used to determine the rotation matrix uniquely except for degenerate cases discussed later. The two vectors represent the orientation of the $z$ and $y$ axes of the bin in the camera coordinates. We denote these vectors as $\vec{v}_z$ and $\vec{v}_y$, respectively.

\begin{figure*}[ht]
  \centering
  \includegraphics[width=\linewidth]{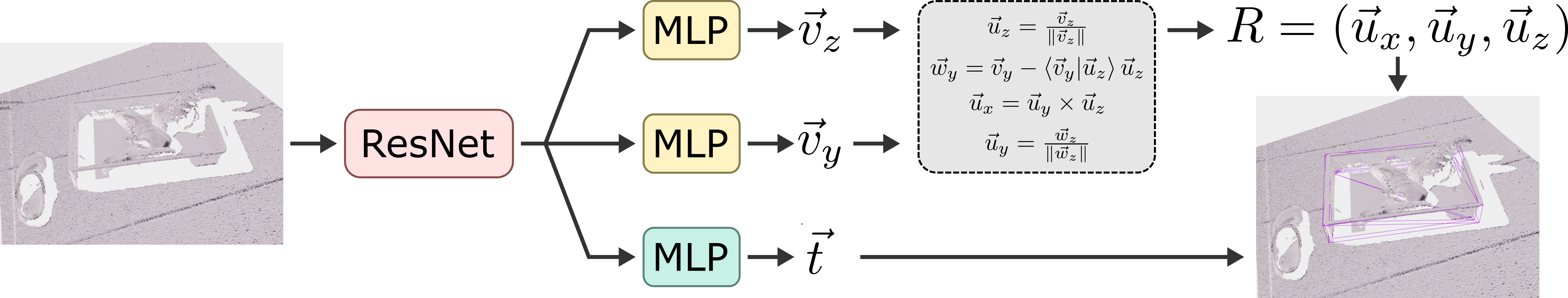}
  \caption{\label{fig:network}%
           The architecture of the bin-pose estimation network. The structured point cloud is fed into a ResNet backbone. The resulting features are fed into three separate heads. Each head is composed of a few fully-connected layers. One of the heads outputs the resulting translation vector $\vec{t}$. The other two heads output intermediate vectors $\vec{v}_z$ and $\vec{u}_z$. Equations (\ref{eqn:base_first}-\ref{eqn:base_cross_prod}) are then used to obtain the columns of the resulting rotation matrix $R$.}
\end{figure*}

To obtain the rotation matrix $R$ from the vectors $\vec{v}_z$ and $\vec{v}_y$, we employ the Gram–Schmidt orthogonalization process to calculate the columns of the actual rotation matrix they represent. During the procedure, we perform the following calculations:
\begin{equation}
\vec{u}_z = \frac{\vec{v}_z}{\lVert\vec{v}_z\rVert},
\label{eqn:base_first}
\end{equation}

\begin{equation}
\vec{w}_y = \vec{v}_y - \left<\vec{v}_y | \vec{u}_z \right> \vec{u}_z,
\end{equation}

\begin{equation}
\vec{u}_y = \frac{\vec{w}_z}{\lVert\vec{w}_z\rVert},
\end{equation}

\begin{equation}
\vec{u}_x = \vec{u}_y \times \vec{u}_z.
\label{eqn:base_cross_prod}
\end{equation}
The vectors $\vec{u}_x, \vec{u}_y, \vec{u}_z$ form an orthonormal basis of $\mathbb{R}^3$. We can then construct a matrix $\left(\vec{u}_x, \vec{u}_y, \vec{u}_z\right)$ which is a valid rotation matrix. The fact that the matrix represents a proper rotation (e.g. $\text{det}(R) = 1$) is enforced by equation $(\ref{eqn:base_cross_prod})$.

Using this procedure, any two vectors $\vec{u}_z$ and $\vec{u}_y$ can yield a valid rotation matrix provided that they are linearly independent. We found this limitation to not be of concern in practice.

Under this parameterization, any rotation matrix can be parameterized by many pairs of such vectors, and it is thus not unique in this regard. However, this is not an issue as we use a loss function which only depends on orientations of $\vec{u}_z$ and $\vec{u}_y$, which are unique. To obtain a single pair of valid vectors $\vec{u}_z$ and $\vec{u}_y$, which would yield a given matrix $R$, we can use the third and second columns of the matrix.
\label{sec:net_param}

\subsubsection{Bin Symmetry}

We aim to detect bins of rectangular shapes. Rectangular bins are symmetric in a 180-degree rotation around an axis parallel to the bin-base normal going through the center of the base. Therefore, there are always two valid rotation matrices for each possible bin pose, which introduces issues during training as the network is forced to learn only one correct output of two possible outputs for a similar input, resulting in the network's inability to converge. 

To remedy this issue we employ a simple strategy. The two possible rotations $R_1$ and $R_2$ are related by a symmetry rotation (\ref{eqn:sym_rot}) such that $R_1 = R_s R_2$, where
\begin{equation}
R_s = \begin{pmatrix} -1 & 0 & 0 \\ 0 & -1 & 0 \\ 0 & 0 & 1 \end{pmatrix}.
\label{eqn:sym_rot}
\end{equation}
Therefore, the only differences between the matrices are the signs in the first two columns, which allows us always to choose one of the matrices based on the sign of the matrix elements. We always select the matrix which has a positive element in the first row and second column. If this element is zero, we use the sign of the next element below. If the value is zero again, we use the sign of the last row and second column, which has to be 1 or -1.

\label{sec:net_symmetry}

\subsubsection{Network Architecture}

In our experiments we use a standard ResNet backbone \cite{he:2016:resnet} for feature extraction. We apply global average pooling on the feature maps and feed the resulting features into three separate branch-heads to output the three vectors $\vec{v}_z, \vec{v}_y$ and $\vec{t}$. Each head comprises two fully-connected layers, with ReLU activations used in rotational heads and Leaky ReLU activations used in the translational branch. The whole network architecture, along with output post-processing, is shown in Figure \ref{fig:network}.

\subsubsection{Loss Function}

For a given ground truth pose defined by $R$ and $\hat{\vec{t}}$ we first check whether to transform the rotation matrix using $R_s$ as described in subsection \ref{sec:net_symmetry}. We extract the vectors $\vec{u}_z$ and $\vec{u}_y$ as the third and second columns of the rotation matrix. We then train the network, which outputs three vectors $\vec{u}_z, \vec{u}_y, \vec{t}$ using a joint loss function:
\begin{equation}
    L = L_{r}(\vec{u}_z, \vec{v}_z) + L_{r}(\vec{u}_y, \vec{v}_y) + \lambda L_{\text{L1}} (\hat{\vec{t}}, \vec{t}),
\end{equation}
where $L_{\text{L1}}$ is the standard L1 loss, $\lambda$ is a weight hyperparameter and $L_r$ is the angle between two vectors in radians:
\begin{equation}
    L_r(\vec{u}, \vec{v}) = \text{acos}\left(\frac{\left<\vec{u} | \vec{v} \right>}{\lVert \vec{u} \rVert \lVert \vec{v} \rVert + \epsilon} \right),
\end{equation}
with $\epsilon$ added to prevent undefined loss for output vectors with small norm.

\section{\uppercase{Evaluation and Final Experiments}}
\label{sec:experiments}

We evaluate the analytical method proposed in Section \ref{sec:analytical} and the neural network described in Section \ref{sec:neural} using the dataset described in Section \ref{section:dataset}. We also show the results after refinement of the network output with ICP and provide an experimental comparison of our method to existing approaches.

\subsection{Evaluation Metrics}

\label{sec:metrics}

Since we do not have 3D surface reconstruction of every bin in our dataset, we rely on model-independent pose error functions, i.e. comparing just the ground truth $\hat{P} = (\hat{R}, \hat{\vec{t}} \,)$ and estimated $P = (R, \vec{t}\,)$ transformation matrices. All our ground-truth rotation matrices consider the same orientation of the cuboid bin with the longer dimension along the $x$-axis, therefore we can use the strategy from subsection \ref{sec:net_symmetry} to obtain symmetries $\hat{R_1}, \hat{R_2}$ and minimize the metrics. We plan to complete the dataset with model reconstructions in the future. This will allow the calculation of metrics like $e_\mathrm{ADI}, e_\mathrm{VSD}, e_\mathrm{MSSD}$ allowing for evaluation of the actual surface alignment \cite{hinterstoier:2012:model}.

Evaluating the translation $\vec{t}$ is straight-forward using the euclidean distance $e_\mathrm{TE}(\, \hat{\vec{t}}, \vec{t} \,) = \lVert\,\vec{t} - \hat{\vec{t}}\,\rVert_2$. For comparison of rotation, we use the angular distance $e_\mathrm{RE}(\hat{R}, R)$, which is the angle between rotational axis in angle-axis representation and can be directly computed from the rotation matrices as: 
\begin{equation}
e_\mathrm{RE}(\hat{R}, R) = \min_{\hat{R'} \in \{\hat{R_1}, \hat{R_2}\}} \mathrm{arcos} \left( \frac{\mathrm{Tr}(\hat{R'}R^{-1})-1}{2} \right),
\end{equation}
where $\mathrm{Tr}$ is the matrix trace operator. 


\subsection{Baseline Network Results}

We have experimented with different configurations of the proposed baseline network\footnote{\url{https://github.com/gajdosech2/bin-detect}}, see Table \ref{tab:results} for results. Apart from the backbone, we tried two different input resolutions, half and quarter of the raw scan, which resulted in resolutions $1032 \times 772$ and $516 \times 386$, respectively. 
ResNet18 with half-resolution of the input scan has the worst performance, probably due to the small receptive field of the network. Interestingly, ResNet34 with quarter-resolution outperformed half-resolution. Additional sub-sampling probably acted as a noise-suppression. 


{\renewcommand{\arraystretch}{1.2}
\begin{table}[t]
\centering
\adjustbox{max width=\columnwidth}{
\begin{tabular}{p{14mm}|c|c|c|c|c|c}
\hline
Backbone & R & S & $L_{r}^z$ & $L_{r}^y$ & $\overline{e_\mathrm{TE}}$ & $\overline{e_\mathrm{RE}}$ 
\\
\hline
\hline
ResNet18 & 1/4 & \ding{51} & 0.058 & 0.198 & 3.808 & 0.256 \\
\hline
ResNet34 & 1/4 & \ding{51} & 0.057 & \textbf{0.145} & \textbf{3.469} & \textbf{0.197} \\
\hline
ResNet18 & 1/2 & \ding{51} & 0.070 & 0.249 & 5.791 & 0.234 \\
\hline
ResNet34 & 1/2 & \ding{51} & 0.063 & 0.222 & 3.979 & 0.266 \\
\hline
\hline
ResNet34 & 1/4 & \ding{55} & \textbf{0.042} & 0.281 & 5.379 &0.323 \\
\hline
\end{tabular}}
\caption{Comparison of test errors of different configurations. Column R denotes the fraction of raw scan resolution used as network input. Column S denotes whether synthetic samples were used during training.}
\label{tab:results}
\end{table}
}

Additionally, we have trained the best performing configuration on a subset of the dataset without the synthetic samples. Naturally, it achieved the worst test error since this set also contains synthetic scans, which were not encountered during training. Surprisingly, it also has higher errors $\overline{e_\mathrm{TE}}=7.656, \overline{e_\mathrm{RE}}=0.559$ on a subset of the test data with real samples only. Configuration trained on both real and synthetic samples achieves $\overline{e_\mathrm{TE}}=6.108$ and $\overline{e_\mathrm{RE}}=0.529$ on such subset. This would suggest that the synthetic data helps the model generalize on real scans, despite the evident gap between real and synthetic samples.  

\begin{figure}[ht]
  \centering
  \includegraphics[width=.99\linewidth]{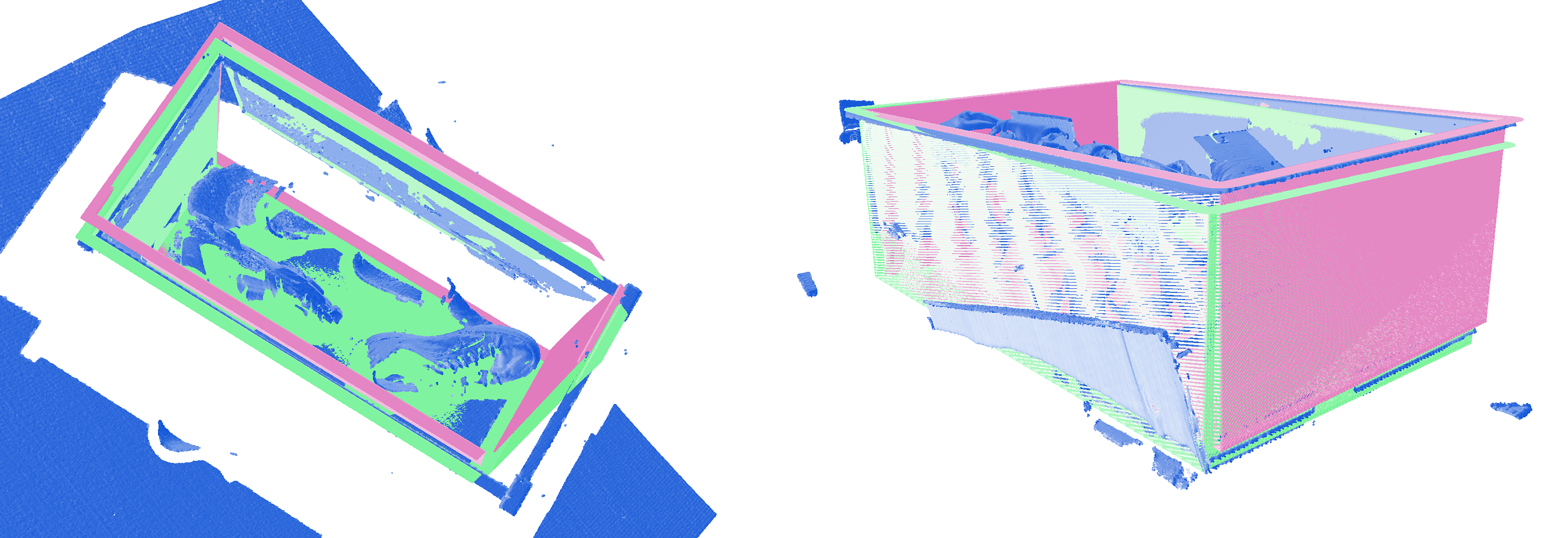}
   \caption{\label{fig:bin_icp}%
            (Left) final improvement of data-driven method using ICP algorithm, (right) a fail-case of the ICP, where the bin was snapped to ground points of the bin, worsening the fit. Points of raw scan are in blue, prediction of the network in pink and ICP refinement in green.}
\end{figure}

Apart from average values of metrics $\overline{e_\mathrm{RE}}, \overline{e_\mathrm{TE}}$, the Table \ref{tab:results} also shows average losses $L_{r}^z = L_{r}(\vec{u}_z, \vec{v}_z)$ and $L_{r}^y = L_{r}(\vec{u}_z, \vec{v}_z)$ over the validation set. 
The loss function has, in this case, useful interpretation even as an evaluation metric.
$L_{r}^z$ represents the error in the predicted normal of the bin's bottom face, with $L_{r}^y$ denoting the rotation around this axis.

\begin{figure*}
  \center
  \includegraphics[width=.94\linewidth]{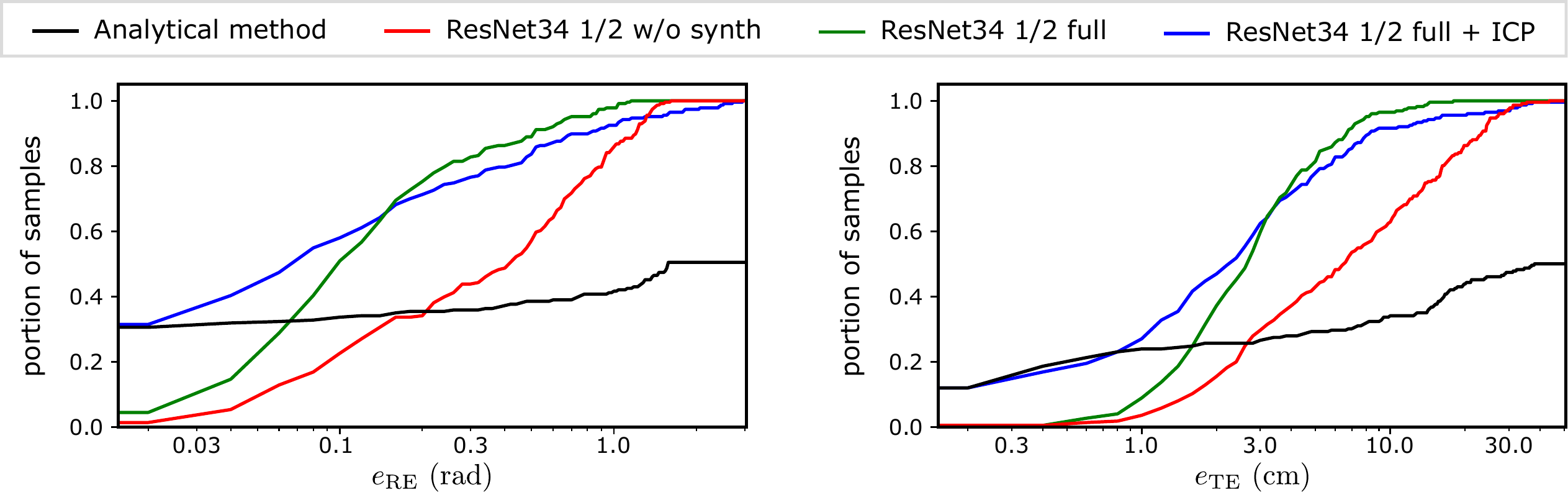}
  \caption{\label{fig:graph}%
    Vertical axes show the fraction of the test samples with the error below the value of the metrics $e_\mathrm{RE}, e_\mathrm{TE}$ on the horizontal axes. The analytical method achieves low error on a few samples but fails to predict pose for approximately half of the cases. Using synthetic data in training improves the overall performance of the neural network. The hybrid method with ICP refinement lowers the minimum error of the network, matching the analytical approach while also retaining robustness. However, in some cases, the ICP fails to improve the bin pose, resulting in slightly increased overall maximum error.
    }
\end{figure*}

A qualitative sample of the hybrid two-step approach, where the data-driven method is refined with post-alignment using ICP, can be seen in Figure \ref{fig:bin_icp}. This refinement improved the results (both $\overline{e_\mathrm{TE}}$ and $\overline{e_\mathrm{RE}}$) in 91 samples out of the 218 in validation + test set. In general, it improves the pose estimation if the bin model has exact size and walls are visible. However, as mentioned in Section \ref{section:dataset}, the dataset currently does not contain complete surface reconstructions of the bins, just their approximate bounding boxes. 

Figure \ref{fig:graph} shows the comparison between the baseline network, its results after ICP refinement, the same version trained over real data only, and our analytical method. 
As can be seen, the analytical method achieves reasonable error for approximately 40\% samples. The remaining samples had either high errors or the method failed to estimate in 47\% cases, which was treated as an infinite error. The ICP refinement achieved almost zero error in a few cases. However, samples with non-corresponding points aligned produced higher errors which can be improved by limiting the usage of ICP only for confident cases, where the number of paired-points is higher than some threshold. This would mitigate the negative effect in a few cases, lowering the average error.

\subsection{Comparison with existing methods}

\begin{figure}[ht]
  \centering
  \includegraphics[width=.94\columnwidth]{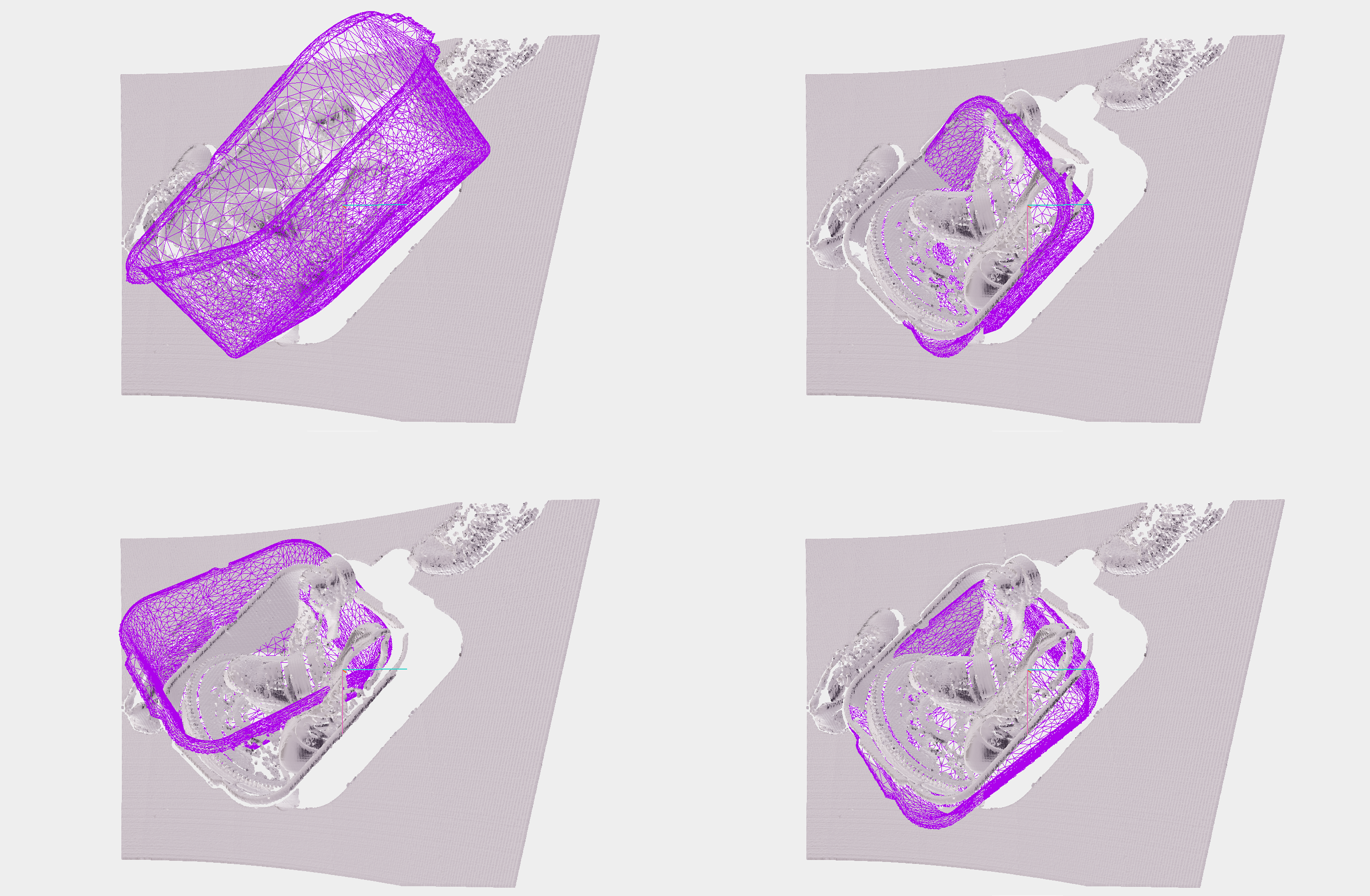}
  \caption{\label{fig:methods}%
    Qualitative comparison on single sample: \textit{Top Left:} DPOD, \textit{Top Right:} EfficientPose, \textit{Bottom Left:} Dense Fusion, \textit{Bottom Right:} Masked Fusion.
    }
\end{figure}

Despite the uniqueness of our data, we have trained and qualitatively evaluated existing state of the art models: DPOD \cite{zakharov:2019:dpod}, DenseFusion \cite{wang:2019:densefusion}, MaskedFusion \cite{pereira:2020:maskedfusion} and EfficientPose \cite{bukschat:2020:efficientpose}. We performed the evaluation only on a subset of our dataset (120 samples) with a single bin model, for which we have made the required surface reconstruction as the compared methods require such data. See Figure \ref{fig:methods} for qualitative comparison 
and Table \ref{tab:comp} for quantitative results over test set of 14 samples. 
We also show the performance of our proposed baseline model. 

{\renewcommand{\arraystretch}{1.3}
\setlength{\tabcolsep}{5pt}
\begin{table}[ht]
\centering
\begin{tabular}{l|r|r|r|r}
\hline
Model & $\overline{e_\mathrm{TE}}$ & $\overline{e_\mathrm{RE}}$ & std $e_\mathrm{TE}$ & std $e_\mathrm{RE}$
\\
\hline
DenseFusion & 7.544 & 0.493 & 2.473 & 0.364 \\
MaskedFusion & 6.583 & 0.494 & 2.145 & 0.361  \\
EfficientPose & 4.148 & 0.454 & 2.256 & 0.308  \\
\hline
Ours & 4.024 & 0.418 & 2.124 & 0.368 \\
\hline
\end{tabular}
\caption{Results over small test set of 14 samples.}
\label{tab:comp}
\end{table}
}

The scope of this experiment is limited, and further evaluation is necessary to draw any strong conclusions. However, this preliminary experiment shows that our method can outperform the existing ones while being conceptually simpler and not requiring a model of the detected bin during training.


\section{\uppercase{Conclusions and Future work}}
\label{sec:conclusion}
In this paper, we have introduced a task of bin pose estimation, which we identified as an essential component in many vision-based automation systems in the industry. We have collected a dataset of high-quality 3D scans of various bins in different environments using scanners with various parameters. In our future work, we aim to improve the dataset by collecting more data to enable a more thorough evaluation of bin pose estimation methods. We hope that such data will be useful for further research in this area.

We also propose two baseline methods for 6D bin pose estimation. The evaluation results suggest that the bin poses can be estimated reliably with a simple convolutional neural network. In many cases, the resulting poses can be further refined using ICP to improve the accuracy of poses. We see the potential for further research in this area, especially regarding the effects of different types of bin pose parametrization on the network performance.


\section*{\uppercase{Acknowledgements}}

The authors gratefully acknowledge the support of NVIDIA Corporation with the donation of GPUs.


\bibliographystyle{apalike}
{\small
\bibliography{example}}

%

\end{document}